\documentclass[runningheads]{llncs}
\usepackage{acronym}
\usepackage{bbm}
\usepackage{multirow}
\usepackage{todonotes}
\usepackage{subfigure}
\usepackage{siunitx}  
\usepackage{graphicx}
\usepackage{hyperref}
\usepackage{amsmath}
\usepackage{rotating}
\usepackage{array}
\usepackage{float}
\usepackage{hhline}
\usepackage{tabularx}
\usepackage[inkscapeformat=png]{svg}

\begin{document}
\title{Standardizing Your Training Process for Human Activity Recognition Models – A Comprehensive Review in the Tunable Factors\thanks{Supported by the Ministry of Science, Research and the Arts Baden-Wuerttemberg as part of the SDSC-BW and by the Carl-Zeiss-Foundation as part of "stay young with robots" (JuBot) project.}}
\titlerunning{Formal HAR}

\author{Yiran Huang\orcidID{0000-0003-3805-1375} \and 
Haibin Zhao\orcidID{0000-0001-7018-1159}\and 
Yexu Zhou\orcidID{0000-0002-8866-7998} \and
Till Riedel\orcidID{0000-0003-4547-1984}\and
Michael Beigl\orcidID{0000-0001-5009-2327}}

\authorrunning{Y. Huang et al.}



%
\institute{Telecooperation Office, Karlsruhe Institute of Technology, Karlsruhe, Germany \email \\{yhuang, hzhao, zhou, riedel, beigl@teco.edu}\\}
%
\maketitle              
\acrodef{har}[HAR]{Human Activity Recognition}
\acrodef{whar}[WHAR]{Wearable Human Activity Recognition}
\acrodef{dnn}[DNN]{Deep Neural Network}
\acrodef{sota}[SOTA]{state-of-the-art}
\acrodef{dl}[DL]{Deep Learning}
\acrodef{ml}[ML]{Machine Learning}
\acrodef{cnn}[CNN]{Convolution Neural Network}
\acrodef{gru}[GRU]{Gated Recurrent Units}
\acrodef{lstm}[LSTM]{Long Short-Term Memory}
\acrodef{loso}[LOSO]{Leave-One-Subject-Out}
\acrodef{cv}[CV]{Cross-Validation}
\acrodef{ai}[AI]{Artificial Intelligence}

\begin{abstract}
In recent years, deep learning has emerged as a potent tool across a multitude of domains, leading to a surge in research pertaining to its application in the \ac{whar} domain.
Despite the rapid development, concerns have been raised about the lack of standardization and consistency in the procedures used for experimental model training, which may affect the reproducibility and reliability of research results. 
In this paper, we provide an exhaustive review of contemporary deep learning research in the field of \ac{whar} and collate information pertaining to the training procedure employed in various studies. Our findings suggest that a major trend is the lack of detail provided by model training protocols. 
Besides, to gain a clearer understanding of the impact of missing descriptions, we utilize a control variables approach to assess the impact of key tunable components (e.g., optimization techniques and early stopping criteria) on the inter-subject generalization capabilities of HAR models. With insights from the analyses, we define a novel integrated training procedure tailored to the \ac{whar} model. Empirical results derived using five well-known \ac{whar} benchmark datasets and three classical HAR model architectures demonstrate the effectiveness of our proposed methodology: in particular, there is a significant improvement in macro F1 \ac{loso} \ac{cv} performance.

\end{abstract}
\newcommand{\vw}{\ensuremath{\mathrm{\boldsymbol{w}}}}
\newcommand{\vb}{\ensuremath{\mathrm{\boldsymbol{b}}}}
\newcommand{\vW}{\ensuremath{\mathrm{\boldsymbol{W}}}}
\newcommand{\vS}{\ensuremath{\mathrm{\boldsymbol{S}}}}
\newcommand{\vx}{\ensuremath{\mathrm{\boldsymbol{x}}}}
\newcommand{\vX}{\ensuremath{\mathrm{\boldsymbol{X}}}}
\newcommand{\vg}{\ensuremath{\mathrm{\boldsymbol{g}}}}
\newcommand{\vc}{\ensuremath{\mathrm{\boldsymbol{c}}}}
\newcommand{\vo}{\ensuremath{\mathrm{\boldsymbol{o}}}}

\section{Introduction}
\label{sec:introduction}

    

Wearable Human Activity Recognition (WHAR) is a process that aims to classify human actions within a specific temporal boundary based on measurements, e.g., acceleration, rotation speed, and geographical coordinates, gathered from personal digital devices~\cite{straczkiewicz2021systematic}. It is central to numerous applications and services. For instance, in the smart homes~\cite{du2019novel}, accurately detecting a user's activity, such as watching television, can enable the automation of derivative tasks like adjusting room light levels. Similarly, in industrial manufacturing~\cite{gunther2019activity}, recognizing a worker's current task could aid in suggesting sequential actions for workflow optimization. In the context of road safety~\cite{pEDAstress}, identifying a driver's present driving behavior or mental states can catalyze measures to improve road safety. These applications place high-performance demands on a \ac{whar} system, necessitating the guarantees of sufficient data and high-performance models.

Wearable technology has experienced significant growth in recent years, with nearly 500 million units shipped globally by the end of 2022, a 500\% increase from the 82 million units shipped in 2016. Smartwatches are projected to be the most popular, accounting for 216 million units in 2022. Smart trackers and glasses follow, with 105 million and 32 million units, respectively. Equipped with various sensors, these devices offer advantages such as unobtrusive continuous data collection, direct user contact, and long battery life. Consequently, they provide detailed insights into human behavior, making wearables a key platform for developing applications that automatically recognize human activity.

According to Google Scholar~\footnote{\url{https://scholar.google.com}.}, from 2021 to the present, a total of 1,660 academic articles have contained the keywords "human activity recognition" and "neural networks" in their titles.

Although these factors have contributed to the rapid growth of \ac{whar}, they also pose challenges in terms of reproducing recent work and benchmarking research. There is a noticeable lack of attention given to explicating the process by which experimental models are trained. The training process of a model, encompassing the trajectory of model development from initialization to gaining expertise in a specific task, plays a pivotal role in determining the model's final performance. 

A comprehensive description of the training process should include the following components: Dataset Description, Model Parameters Setting if present, Data Preprocessing, Batch Size and Maximum Epoch Setting, Optimizer Setting, Learning rate strategy, Early stopping strategy, Final model selection strategy, Validation\&Test setting, Additional details.

By comprehensively articulating each stage of the training process, the intention of the training process description is to provide the reader with the ability to accurately replicate the experimental result described in the publication. Such documentation reinforces the transparency and integrity of the research, elements that are paramount in the field of deep learning science.

Nonetheless, our analysis points out that while nearly all publications deliver comprehensive details about dataset descriptions, model parameters setting, data preprocessing, and validation and test settings, there is a notable lack of exhaustive detail in other aspects.

Until now, the majority of review articles in the realm of \ac{whar} have mainly concentrated on outlining various machine learning and deep learning architectures, with an emphasis on model design. Contrary to this trend, our research pivots its focus towards a comprehensive analysis of the model training process in the context of \ac{whar}. To the best of our knowledge, this survey stands as a pioneering effort specifically targeting descriptions of the model training process. Through this exploration, we endeavor to shed light on the current state of model training process narratives in the relevant literature and proffer recommendations to guide future scholarly pursuits in this domain.

The key contributions of our work can be summarized below:
\begin{itemize}
    \item To the best of our knowledge, we are the first in the \ac{whar} field to write a review that focuses on the model training procedure in various research papers. 
    \item This study delivers insights into how models have been trained in recent years. It identifies prevailing issues about model training in the field and highlights the potential consequences of these problems using control variables.
    \item In response to these identified issues, we propose a countermeasure and provide experimental evidence of its efficacy.
    \item We are releasing the proposed training procedure as an open-source package to aid in further research in this field.
\end{itemize}


\begin{table}[]
    \centering
    \caption{Summarize of the hyperparameters and the corresponding range in model training procedure.}
    \begin{tabularx}{\textwidth}{|c|X|}
        \hhline{|==|}
        Parameter& Range  \\
        \hline
        Optimizer&\{sGD~\cite{robbins1951stochastic}, ADAM~\cite{kingma2014adam}, Adadelta~\cite{zeiler2012adadelta}, Rmsprop\}\\ 
        \hline
        Learning Rate&[$1e-1$, $1e-5$]\\
        \hline
        Weight Decay&[$1e-1$, $1e-8$]\\
        \hline
        Scheduler&\{Step~\cite{tang2022multiscale}, LRPlateau~\cite{simonyan2014very}, Cos~\cite{loshchilov2016sgdr}, Cos+restart~\cite{loshchilov2016sgdr}\}\\
        \hline
        Scheduler Base&\{Train Loss, Validation Loss, Train Metric, Validation Metric\}\\
        \hline
        Scheduler Patience (epoch)&[1, 100]\\
        \hline
        Batch Size&[16, 512]\\
        \hline
        Maximum Epoch&[10, 1500]\\
        \hline
        Early Stopping&\{Yes, No\}\\
        \hline
        Early Stopping Base&\{Train Loss, Validation Loss, Train Metric, Validation Metric\}\\
        \hline
        Early Stopping Patience (epoch)&[1,100]\\
        \hline
        Model Selection Base&\{Last, Train Loss, Validation Loss, Train Metric, Validation Metric\}\\
        \hline
    \end{tabularx}
    \label{tab:hyperparameter}
    \vspace{-10pt}
\end{table}

\section{Search Strategy}
\label{sec: search strategy}
We followed Preferred Reporting Items for Systematic review and Meta-Analysis Protocols (PRISMA-P) to select relevant and important articles for our research target. We completed this search using two protocols: the search protocol and the three-level exclusion protocol.

\textbf{Search Protocol:} We utilized Google Scholar, a free academic search engine provided by Google, as our search platform. It provides an extensive coverage of academic literature across numerous disciplines worldwide, such as journal articles, theses, conference papers, and more. Its database is regularly updated, ensuring access to the most recent academic literature. Google Scholar supports keyword-based searches to retrieve related articles. In alignment with our research goals, we initially used the following three keyword sets for our exploration: ("Human Activity Recognition" OR "HAR"), ("Machine learning" OR "Deep learning" OR "Neural network"), and ("Wearable" OR "Smartphone" OR "Smartwatch" OR "Smartglasses" OR "Smartband"), combined with the platform's limited search statements. In light of our objective to analyze the latest developments, we limited the search time frame to 2021 through June 2023. Through this procedure, we sourced a total of 4,960 articles.

\textbf{Three-Level Exclusion Protocol:} We prioritized relevance and ranked papers by their impact and subsequently excluded duplicates, non-English articles, non-conference/journal papers, and those lacking references. We filtered out articles whose titles did not align with our research purpose. This initial inspection helped us narrow down to 274 articles. By reading the abstracts of these 274 articles, we removed survey articles and those that focused on other types of \ac{har} tasks. The second filtering phase yielded 208 papers. On downloading these papers and swiftly reviewing their contents, we discarded those primarily focusing on machine learning models, data collection, and data processing. The third and final filtering round culminated in a total of 102 papers.


\section{Observation}
\label{sec: observation}
In deep learning, beyond the inherent model parameters optimized using gradient descent, several hyperparameters critically influence a model's performance. These include the number of layers, kernel sizes, and training aspects like optimizer selection and batch size. These hyperparameters are not directly learned through training but profoundly affect the model's structural complexity, training speed, and overall efficacy.

The description of the training process in a research paper is essentially a record of the values taken for these parameters. It ensures that the entire training process of the model is unambiguous. ~\autoref{tab:hyperparameter} summarizes the hyperparameters that describe the training process, along with their common ranges within the \ac{whar} domain as derived from 102 studied papers.

For a specific task, hyperparameter tuning aims to identify the optimal configuration of hyperparameters leading to the best model performance e.g. through Bayesian Optimization~\cite{wu2019hyperparameter}, reinforcement learning~\cite{huang2022universal} or Monte Carlo~\cite{huang2022automatic,huang2023mcxai}. However, as demonstrated in ~\autoref{tab:hyperparameter}, the search space for training model hyperparameters is vast. It leads to substantial computational costs associated with hyperparameter tuning, especially for tasks like \ac{whar} that involve large datasets and complex feature extraction using complex models. Individual model training on the classic oppo~\cite{data:Oppo} dataset can take approximately one hour on Tesla P100~\footnote{The performance is tested with \url{https://colab.research.google.com/}.}. Performing systematic hyperparameter searches incurs significant time and computational resource expenses. Furthermore, considering typical limitations in paper length and the focus of \ac{whar}-related papers on introducing new model structures, algorithms, or theories, hyperparameter tuning is generally not the primary focus. Among the 102 papers we analyzed, only 16 of them (15.7\%) involved hyperparameter tuning. 

We observe that all these works only explored a subset of the hyperparameter space. Most concentrated on seeking for the learning rate and batch size, followed by the number of epochs and the choice of optimizer. There is a lack of exploration in other hyperparameter dimensions. Additionally, we found that the results of the search efforts were relatively consistent regarding optimizer and learning rate. All searches pertaining to the optimizer ultimately converged on selecting the Adam optimizer~\cite{kingma2014adam}. The majority of chosen learning rates were around $1e-3$, accounting for $64.3\%$ of similar searches. Epoch values mostly centered around $60$. Batch size results were polarized, with half of the studies selecting a size of $64$, while the other half opted for batch sizes above $400$. This apparent inconsistency in the search results further emphasizes the significance of hyperparameters. It underscores that the required hyperparameters can vary significantly based on differences in models and data.

\begin{figure}
    \centering
    \includegraphics[width=0.9\linewidth]{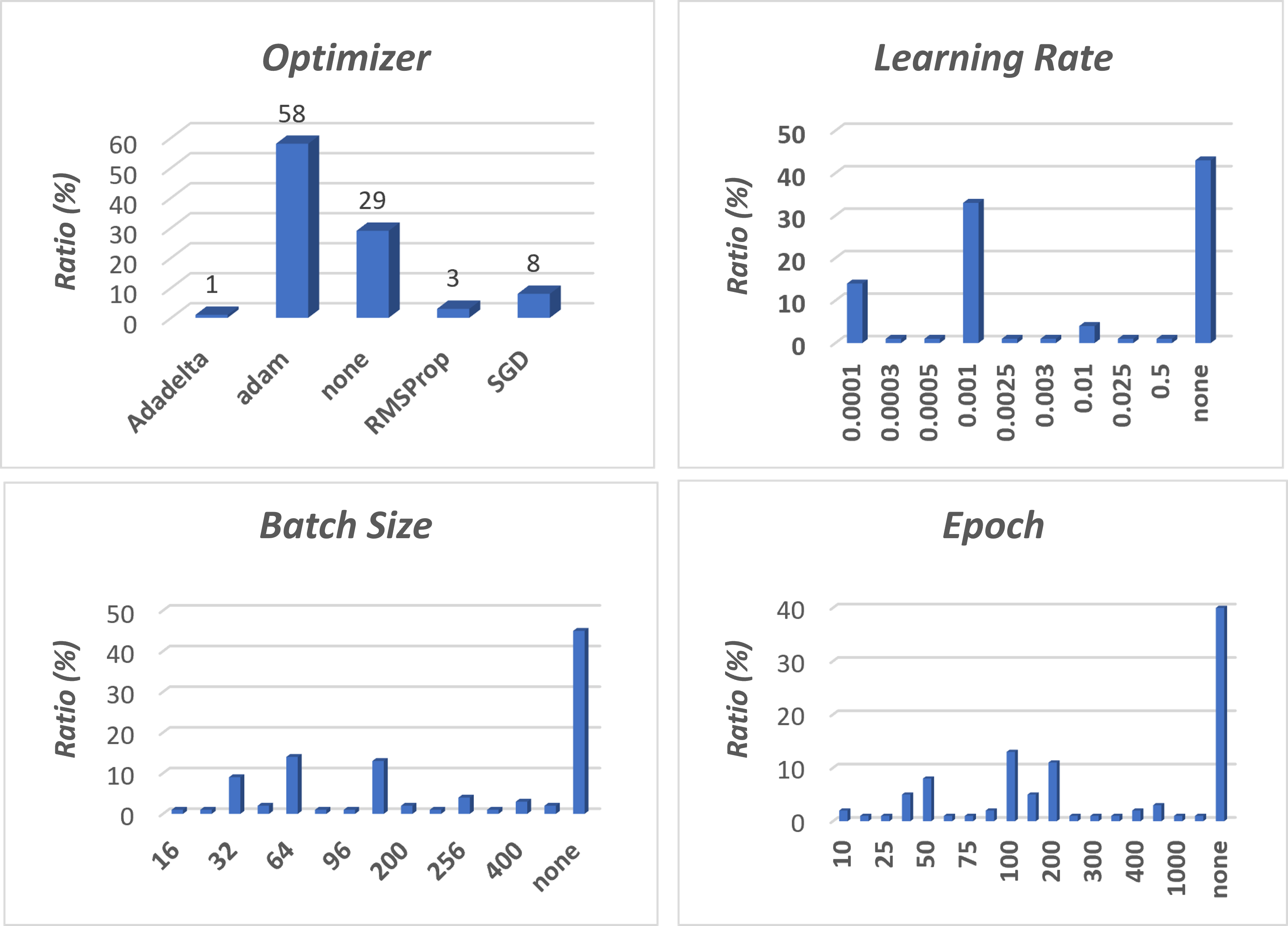}
    \caption{Distribution on the four hyperparameters Optimizer, Learning Rate, Batch Size, and Epoch across all reviewed papers. 'none' indicates that there is no corresponding description in the paper.}
    \label{fig:para distribution1}
    \vspace{-10pt}
\end{figure}

Apart from the $16$ articles that engaged in hyperparameter tuning, the remaining $86$ articles addressed the potential adverse effects of neglecting hyperparameter tuning through various strategies. Some adopted parameter settings from similar works, while the majority rely on empirical choices~\cite{zhao2022improving,huang2023randomhar}. ~\autoref{fig:para distribution1} illustrates the hyperparameter setting made by these articles in terms of the four mentioned parameters. Similar to the tuning results, these studies also predominantly favored the Adam optimizer~\cite{kingma2014adam}. Concerning learning rates, in addition to $1e-3$, $1e-4$ was also a popular choice. The setting of the epoch number deviated from the tuning results. Most of the works leaned towards larger epoch values. This trend could be attributed to the application of techniques like early stopping. In early stopping scenarios, even if a large number of epochs is set, training can be terminated ahead of time if specific conditions are met. For instance, if the validation loss ceases to decrease, training can halt prematurely. These strategies aim to mitigate the negative impact of neglecting exhaustive hyperparameter optimization. While they might not yield the optimal performance, they offer a practical approach given the resource constraints, limited paper space, and the specific focuses of the papers in the \ac{whar} domain.

Additionally, we observed in ~\autoref{fig:para distribution1} that the "none" category has a high occurrence rate across all plots. This signifies that, using batch size as an example, in these papers, 45\% of them did not provide a description of the batch size. This indicates an incompleteness in the descriptions of their model training process. This scenario is likewise seen in other hyperparameter plots.

\begin{figure}
    \vspace{-10pt}
    \centering
    \includegraphics[width=0.6\textwidth]{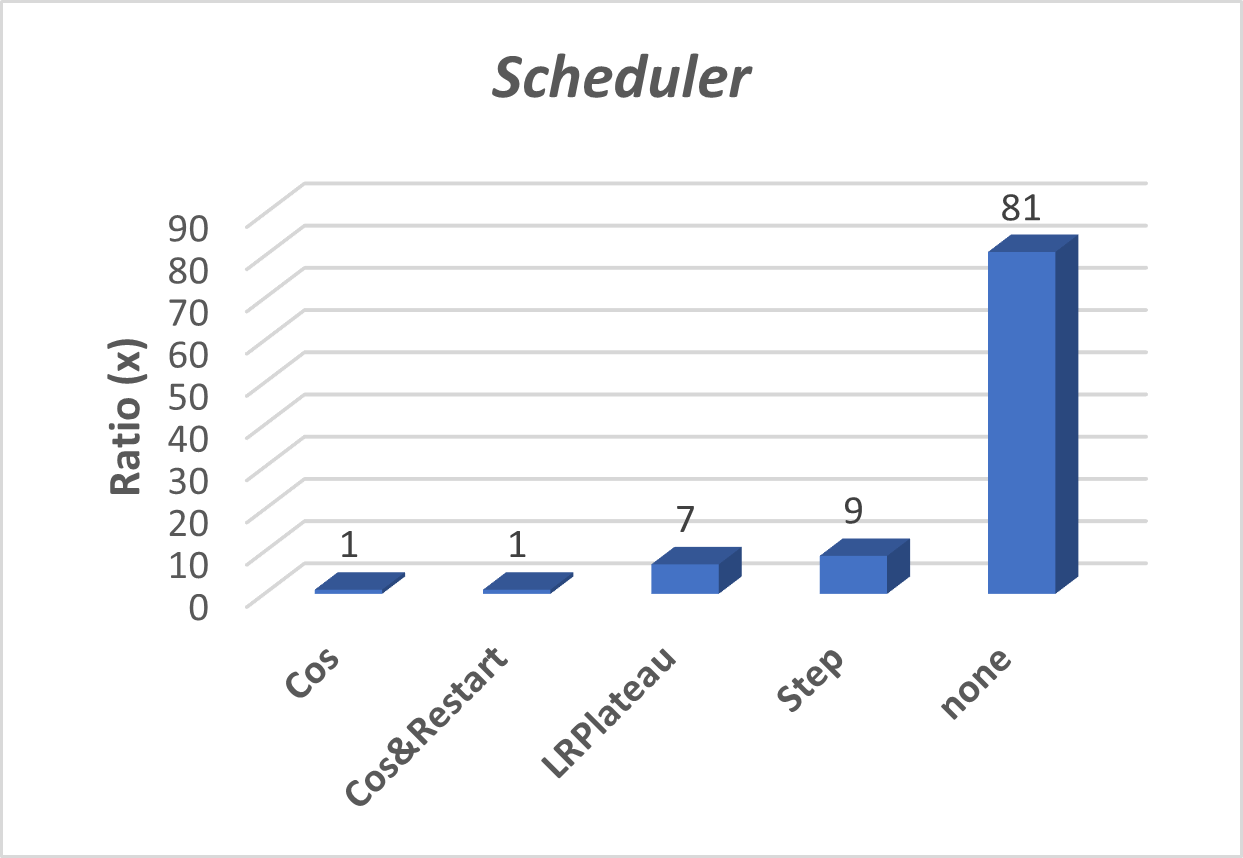}  
    \caption{Distribution of choices for scheduler hyperparameters across all reviewed papers.}
    \label{fig:para distribution2}
    \vspace{-15pt}
\end{figure}

~\autoref{fig:para distribution2} presents a summary of the descriptions related to parameter schedulers across all reviewed papers. It was found that 81\% of the articles lacked descriptions about schedulers. Moreover, only 6\% of the articles described early stopping procedures, and only 9\% outlined the final model selection process. 

Intuitively, it is clear that challenges due to descriptive deficiencies can significantly affect the reliability and reproducibility of the corresponding articles, underscoring the need for a standardized and comprehensive description of model training. 
However, the actual strength of the impact of the parameters describing the model training process, for instance, optimizer settings (type, learning rate, weight decay), early stopping strategies and model selection strategies, on the \ac{whar} model end performance remains an open question, which is discussed in the next section.


\section{Control-Variates Analysis}
\label{sec: control-variates analysis}
\begin{figure*}
    \centering
    \includegraphics[width=.9\textwidth]{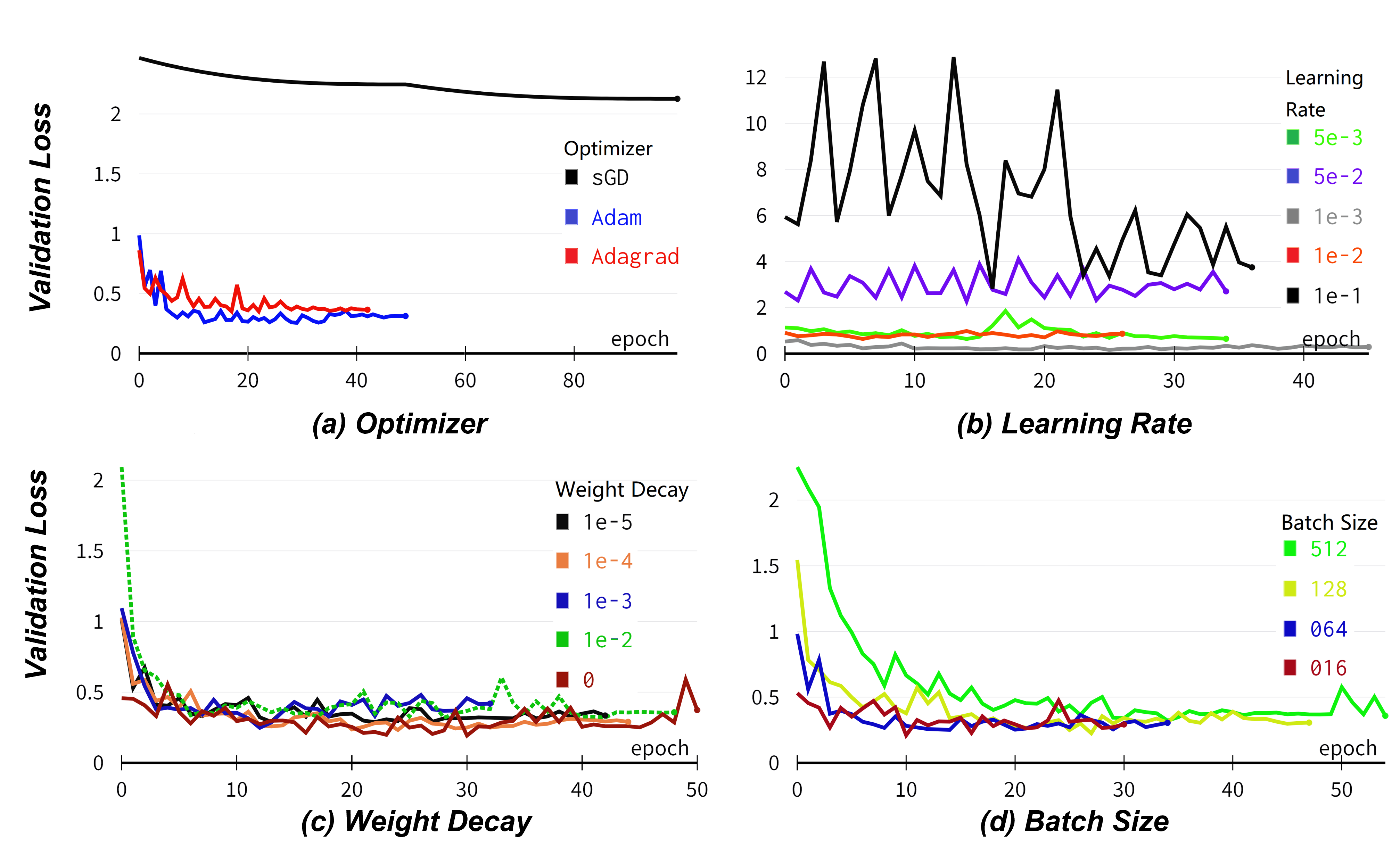}
    \vspace{-5pt}
    \caption{Validation loss during model training in control-variates experiment.}
    \label{fig: cvresult}
    \vspace{-10pt}
\end{figure*}

This section aims to investigate the independent impact of the aforementioned factors on the final model performance in \ac{whar} context. To achieve this, we employ the control-variates method and run experiments with CNNLSTM~\cite{mutegeki2020cnn} model on the HAPT~\cite{data:hapt} dataset. The baseline training procedure is derived from the collected papers by identifying the most frequently utilized hyperparameters (as described in~\autoref{sec: observation}) in the training procedure. Specifically, an Adam optimizer~\cite{kingma2014adam} is employed with an initial learning rate set at $1e-3$ and no weight decay. Cross-entropy is selected as the loss function, along with batch training set to a batch size of $64$. The maximum number of epochs is capped at $60$. A step scheduler is also utilized to reduce the learning rate by a factor of $0.1$ every $10$ epochs. For early stopping, we use a $10$-epoch patience also based on validation loss. The test model selected runs on the least validation loss post-training. \autoref{fig: cvresult} visualizes the validation loss during the training in control-variates experiment for hyperparameter: optimizer, learning rate, weight decay and batch size.

\autoref{fig: cvresult}(a) depicts the impact of diverse optimizers (i.e., sGD~\cite{robbins1951stochastic}, Adagrad~\cite{duchi2011adaptive}, and Adam~\cite{kingma2014adam}) on model training. Our analysis finds notable variance in results across these optimizers. Of the three, Adam~\cite{kingma2014adam} yields the best performance. Conversely, sGD~\cite{robbins1951stochastic} does not converge due to its hyperparameter sensitivity and inappropriate initialization.

\autoref{fig: cvresult}(b) illustrates the effect of different initial training rates. Notably, although the baseline Adam optimizer~\cite{kingma2014adam} can adaptively adjust the step size, which is almost equivalent to adjusting the learning rate, the initial learning rate still plays a crucial role in the model training. In particular, setting an improper initial learning rate may hinder the model from converging (e.g., the black curve), leading to suboptimal performance. In addition, the best model performance is achieved by setting the initial learning rate $1e-3$.

Despite the strong capability of weight decay for improving model generalization in other domains~\cite{lecun1998gradient,hinton2006reducing}, \autoref{fig: cvresult}(c) draws a different conclusion in \ac{whar} domain. We can observe that the utilization of different weight decays does not significantly influence the validation loss. An excessive weight decay may even increase the validation loss, conversely, setting the weight to $1e-4$ seems to facilitate a more stable training process.

\autoref{fig: cvresult}(d) summarizes the validation loss with different batch sizes. Consistent with the conclusion from another independent work~\cite{keskar2016large}, a smaller batch size leads to the better model generalization. However, as reported in~\autoref{fig:para distribution1}, it is not a common option in recent \ac{whar} articles.

\begin{figure}[h]
\vspace{-10pt}
    \centering
    \includegraphics[width=.9\linewidth]{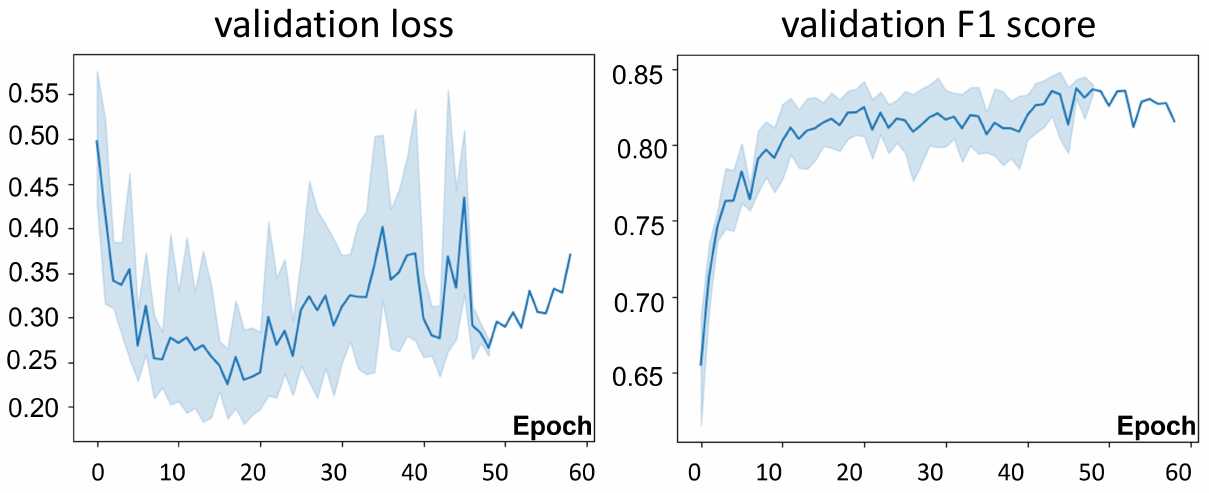}
    \vspace{-10pt}
    \caption{Mean and standard deviation of the validation loss (left) and macro F1 score (right) of the models trained on HAPT dataset.}
    \label{fig:loss&f1score}
    \vspace{-10pt}
\end{figure}

\autoref{fig:loss&f1score} depicts the validation loss and macro F1 score over the epoch during the training on HAPT~\cite{data:hapt} dataset. The solid line represents the mean F1 score w.r.t. subjects, while the shading indicates the corresponding variance. Our analysis reveals that the validation loss decreases gradually and reaches its minimum at approximately the 15th epoch. However, the loss tends to increase progressively thereafter. In contrast, the F1 score continues to improve gradually even after the 15th epoch. This implies that employing validation loss for early-stopping may hinder the model from reaching a higher F1 score, which is the actual metric for evaluating the model performance. Additionally, the model with early-stopping does not perform as well as the model obtained by terminating the training at the maximal training epochs.

Based on the analysis of the control-variates approach, we can conclude that all the aforementioned factors have a discernible impact on the final performance of \ac{whar} models. Therefore, their inclusion in academic papers is crucial to ensuring the reproducibility of experiments.

\begin{table*}[htbp]
    \vspace{-10pt}
    \centering
    \caption{Summary of the datasets used in the experiment. The abbreviations acc, gyro, mag denote 3d accelerometers, gyroscopes and magnetometers, respectively.}
    \begin{tabular}{|c|c|c|c|c|c|c|}
        \hline
         Name & \#Subjects  &\#Channels&\#Length& \#Classes& Sensors type & Freq\\
         \hline
         DSADS & 8 & 45& 126&19 & acc, gyro, mag&$25\,\si{\hertz}$\\
         \hline
         HAPT&30&6&128&12&acc, gyro&$50\,\si{\hertz}$\\
         \hline
         OPPO&4&77&30&18&acc, gyro, mag&$30\,\si{\hertz}$\\
         \hline
         PAMAP2&9&18&168&12&acc, gyro&$100\,\si{\hertz}$\\
         \hline
         RWHAR&15&21&128&8&acc&$50\,\si{\hertz}$\\
         \hline
         
    \end{tabular}
    \vspace{-25pt}
    \label{tab: datasets}
\end{table*}

\section{Experiment}
\label{sec: experiment}
It has been observed that frequently used training hyperparameters derived from reviewed articles do not yield an optimal \ac{whar} model. Accordingly, without using hyperparameter tuning, how should we set the parameters of the training process to get relatively close to optimal results? To address this, we propose a hypothesis: by employing the optimal setting for each factor (determined during the control-variates analysis) for model training (denoted as \textit{new}), the performance of the trained model could be enhanced, surpassing the effectiveness of the baseline used in \autoref{sec: control-variates analysis} (denoted as \textit{comm}). Specifically, our approach involves using an Adam optimizer~\cite{kingma2014adam} with an initial learning rate of $1e-3$
and a weight decay of $1e-4$. In addition, a Cosine Scheduler~\cite{loshchilov2016sgdr} is employed, along with the application of the cross-entropy loss function~\cite{shannon2001mathematical}. We utilize batch training with a batch size of $16$ and apply the F1 score on the validation set for early stopping, with a patience of $30$ epochs. After training, the optimal model is chosen based on the best F1 score on the validation set. In this section, we conduct an experiment to test this hypothesis.


\textbf{Benchmark \ac{whar} datasets.} To ensure experimental fairness, we conducted experiments on five most widely used benchmark datasets, i.e., DSADS~\cite{data:dsads}, HAPT~\cite{data:hapt}, OPPO~\cite{data:Oppo}, PAMAP2~\cite{data:pamap2}, and RW~\cite{data:rw}. As summarized in~\autoref{tab: datasets}, these datasets differ significantly in terms of sensor types, sampling rates, number of subjects, number of channels, sequence length and application scenarios. As data preprocessing, we applied sliding windows to split data in segments, specifically, we employed a $50\%$ overlap between adjacent windows for the training and validation sets. While for the test set, we used a $90\%$ overlap between adjacent windows~\cite{jordao2018human}. 

\textbf{Model} To ensure the generalizability of our new training process, three models with different structures are utilized as our target models, namely, \ac{cnn}-based model (MCNN)~\cite{krizhevsky2017imagenet}, hybrid CNN-LSTM model (CNNLSTM)~\cite{mutegeki2020cnn}, and transformer-based model (Transformer)~\cite{zhang2023human}. 

\textbf{Evaluation and Metric} To evaluate the performance of the final obtained models and demonstrate their generalizability across subjects, we employ the \ac{loso}-\ac{cv} method on all datasets. In response to the category imbalance present within the \ac{whar} data, we apply the F1 score as the evaluation metric. Moreover, to ensure the robustness of our results, we repeat the experiment five times using different random seeds and report the average results for all experiments. 

\textbf{Baseline} We identified the most frequently utilized training procedure in the collected academic papers and use them as our baseline (comm). Concretely, we use an Adam optimizer~\cite{kingma2014adam} with an initial learning rate of $1e-3$ without weight decay. We use the cross-entropy loss to guide the batch training with a batch size of $256$. We employ an LRscheduler that reduces the learning rate by a factor of $0.1$ based on the validation loss with patience of $10$ epochs. We apply early-stopping with the patience of $30$ epochs also on validation loss. Finally, we consider the model with the lowest validation loss as the optimal model.


\subsection{Result}
\autoref{tab: exp} presents the results of our experiments. We can observe that, the new training procedure achieved significantly better results compared to the baseline procedure on all datasets and across all models. This enhancement confirms the effectiveness of the proposed training procedure. In addition, it is noteworthy to mention that, this enhancement can be gained without any supplementary demands on the training resource and any increasing model inference time.

\begin{table*}[htbp]
\vspace{-10pt}
\centering
\caption{\ac{loso}-\ac{cv} performance (mean F1 score) comparison.}
\small
\begin{tabular}{|c|c|c|c|c|c|c|}
\hline
\textbf{Model}& \textbf{Method} & \textbf{DSADS} & \textbf{HAPT} & \textbf{OPPO} & \textbf{PAMAP2} & \textbf{RW}\\
\hline
\multirow{2}{*}{\textbf{MCNN}}&comm&0.801&0.779&0.405&0.708&0.691\\
\cline{2-7}
&new&\textbf{0.865}&\textbf{0.802}&\textbf{0.432}&\textbf{0.734}&\textbf{0.747}\\
\hline
\multirow{2}{*}{\textbf{CNNLSTM}}&comm&0.854&0.806&0.392&0.735&0.706\\
\cline{2-7}
&new&\textbf{0.900}&\textbf{0.820}&\textbf{0.395}&\textbf{0.764}&\textbf{0.768}\\
\hline
\multirow{2}{*}{\textbf{Transformer}}&comm&0.812&0.767&0.402&0.654&0.598\\
\cline{2-7}
&new&\textbf{0.872}&\textbf{0.805}&\textbf{0.440}&\textbf{0.758}&\textbf{0.731}\\
\hline
\end{tabular}
\label{tab: exp}
\vspace{-25pt}
\end{table*}

\section{Conclusion and future work}
\label{sec: conclusion}
In this study, we highlight the prevalent trend that, the majority of academic papers are deficient in providing a comprehensive description of the model training process. Employing the control-variables analysis, we affirm that, these overlooked factors exert a significant influence on the model performance. Therefore, the lack of these descriptions can markedly affect the reproducibility of the achievements reported in those academic papers.
In addition, based on the control-variates analysis, we propose a new training procedure and test its effectiveness by conducting extensive experiments on five widely used benchmark datasets using three models with different architectures. The experiments indicate that, the new training procedure can achieve significantly better results compared to the most common training procedure on all datasets and all models.

%
%
%
\bibliographystyle{splncs04}
\bibliography{references}

\end{document}